\documentclass[sigconf, nonacm]{acmart}

\usepackage{booktabs} 
\usepackage{graphicx}
\usepackage{subfigure}

\usepackage{amsmath,amssymb,amsfonts}
\usepackage{algorithm}
\usepackage{algorithmic}
\usepackage{color}
\usepackage{hyperref}
\usepackage{url}
\usepackage{tabularx}
\usepackage{multicol}
\usepackage{multirow}

\newcommand{\tabularxmulticolumncentered}[3] 
    {\multicolumn{#1}
                 {>{\centering\hsize=\dimexpr#1\hsize+#1\tabcolsep+\arrayrulewidth\relax}#2}
                 {#3}}

\renewcommand{\thefootnote}{\fnsymbol{footnote}}

\hypersetup{
    colorlinks=true,
    linkcolor=blue,
    filecolor=blue,      
    urlcolor=blue,
    citecolor=cyan,
}

\newcommand\vldbdoi{XX.XX/XXX.XX}

\newcommand\vldbavailabilityurl{URL_TO_YOUR_ARTIFACTS}

\begin{document}
\title{Differentiable Voxelization and Mesh Morphing}

\author{Yihao Luo $^*$}
\orcid{0000-0002-1825-0097}
\affiliation{%
  \institution{Imperial College London, UK}
  \streetaddress{}
  \city{}
  \state{}
  \postcode{}
}
\email{y.luo23@Imperial.ac.uk}

\author{Yikai Wang}
\orcid{0000-0002-1825-0097}
\affiliation{%
  \institution{Tsinghua University, China}
  \streetaddress{}
  \city{}
  \country{}
}
\email{yikaiw@outlook.com}

\author{Zhengrui Xiang}
\orcid{0000-0001-5109-3700}
\affiliation{%
  \institution{Imperial College London, UK}
  \city{}
  \country{}
}
\email{z.xiang21@imperial.ac.uk}

\author{Yuliang Xiu}
\orcid{0000-0002-1825-0097}
\affiliation{%
  \institution{Max Planck Institute for Intelligent Systems, Germany}
  \streetaddress{}
  \city{}
  \country{}
}
\email{yuliang.xiu@tuebingen.mpg.de}

\author{Guang Yang$^\dagger$}
\orcid{0000-0001-5109-3700}
\affiliation{%
  \institution{Imperial College London, UK}
  \city{}
  \country{}
}
\email{g.yang@imperial.ac.uk}

\author{ChoonHwai Yap$^\dagger$}
\orcid{0000-0001-5109-3700}
\affiliation{%
  \institution{Imperial College London, UK}
  \city{}
  \country{}
}
\email{c.yap@imperial.ac.uk}

\begin{abstract}
In this paper, we propose the differentiable voxelization of 3D meshes via the winding number and solid angles. The proposed approach achieves fast, flexible, and accurate voxelization of 3D meshes, admitting the computation of gradients with respect to the input mesh and GPU acceleration. We further demonstrate the application of the proposed voxelization in mesh morphing, where the voxelized mesh is deformed by a neural network. The proposed method is evaluated on the ShapeNet dataset and achieves state-of-the-art performance in terms of both accuracy and efficiency. The code is available at \href{https://github.com/Luo-Yihao/DOPH}{\textcolor{blue}{https://github.com/Luo-Yihao/DOPH}}.
\end{abstract}

\maketitle

\begingroup
\renewcommand\thefootnote{}\footnote{\noindent
This work is licensed under the Creative Commons BY-NC-ND 4.0 International License. Visit \url{https://creativecommons.org/licenses/by-nc-nd/4.0/} to view a copy of this license. For any use beyond those covered by this license, obtain permission by emailing \href{mailto:yihaoluo2020@gmail.com}{yihaoluo2020@gmail.com}. Copyright is held by the owner/author(s). \\
$*$ Corresponding author.\\
$\dagger$ Co-last Senior authors.
}\addtocounter{footnote}{-1}\endgroup



\section{Introduction}

\begin{figure*}[!htp]
  \centering
  \includegraphics[width=\linewidth]{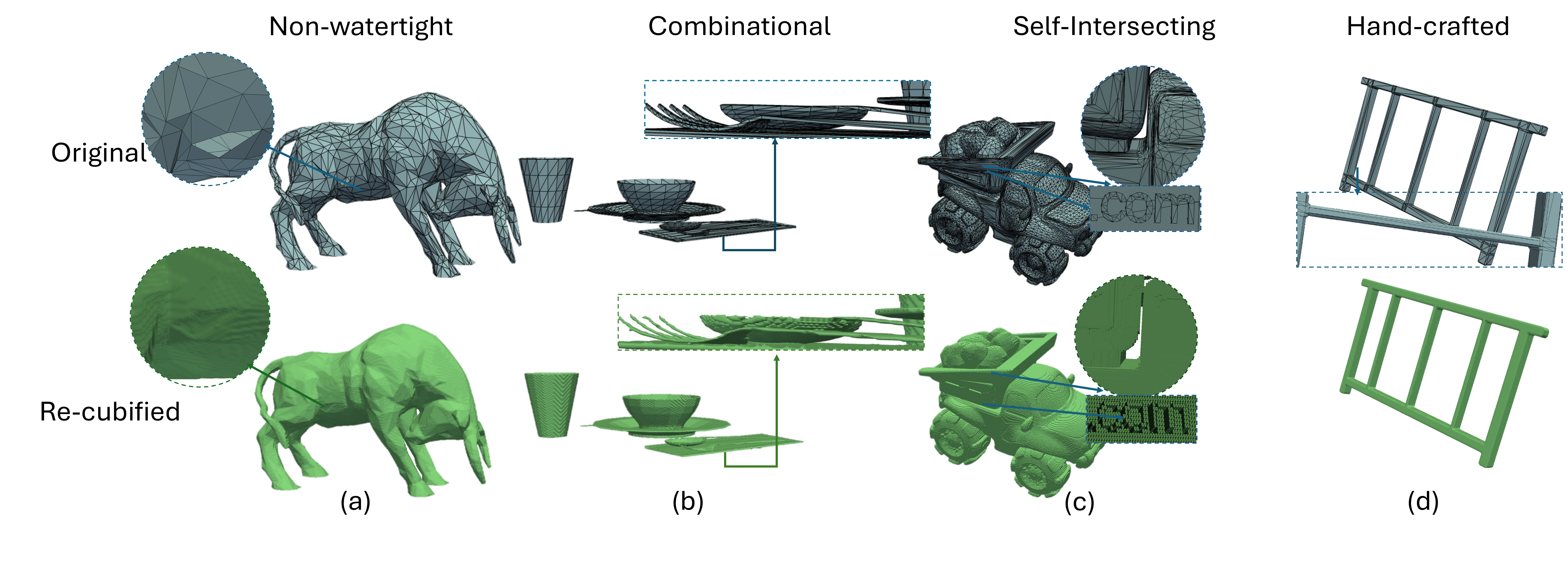}
  \caption{Differentiable voxelization algorithm converts a 3D mesh into a 3D grid of occupancy voxels by the winding number and solid angles on GPU. The first row shows hard cases of the voxelization of 3D meshes using traditional methods, where the mesh is not watertight, contains irregular triangles, or is formed by combinations of shapes or with complex geometry. The second row shows the voxelization results of the proposed differentiable voxelization algorithm, followed by the re-cubifying.}
  \label{fig:diffoccupancy}
\end{figure*}

Voxel is a fundamental representation of 3D objects in computer graphics and computer vision. It is widely used in 3D reconstruction \cite{brock2016generative, Wang_2022DOGN}, 3D rendering \cite{kato2020differentiable, loper2014opendr, liu2019soft}, and 3D generative models \cite{mescheder2019occupancy, threestudio2023,ren2024xcube} due to its uniform grid structure and simplicity. Therefore, voxelization, which converts a 3D mesh into a 3D grid of voxels, is an essential preprocessing step for many 3D deep learning tasks. However, the commonly used voxelization algorithms, such as the rasterization-based and the mesh-boolean-based methods \cite{wang2010approximate,zhou2016mesh, Wang_2022DOGN}, implemented on the CPU, are not efficient and accurate, which consumes a lot of time and memory and limits the development of high-resolution 3D deep learning works. To address this issue, we propose a novel differentiable voxelization algorithm for 3D meshes, which is fast by GPU acceleration and flexible in terms of arbitrary sampling resolution on non-watertight meshes with arbitrary topologies. More importantly, the proposed algorithm is differentiable, which enables the computation of gradients with respect to the input mesh, facilitating the optimization of mesh-level deep learning tasks. The proposed approach is based on the winding number and solid angles. 

In recent work, Luo et al. \cite{luo2023ghd} used the abstract inverse quadratic field to compute the occupancy \cite{mescheder2019occupancy} of voxels and slice a mesh differentiablely, which facilitates the slicing and voxelization of 3D meshes and achieves mesh reconstruction from 2D medical scan images. However, for general 3D meshes with irregular topologies and wiring structures, the abstract inverse quadratic field is not accurate enough due to the integration error. This issue is addressed by replacing field integration with solid angles, which are more accurate and efficient in computing the occupancy of voxels without loss of differentiability. Jacobson et al. \cite{jacobson2013robust, jacobson2016boolean} achieved the voxelization and boolean operations of 3D meshes by the winding number and solid angles, but the naive computation of the solid angles leads to a computational gap near surface boundaries, which requires further human interventions for discretization and smoothing. Furthermore, we discard the hierarchical winding number evaluation from \cite{jacobson2013robust} but implement the solid angles computation in a parallel manner on GPU, which is more efficient and accurate. Leveraging the proposed differentiable voxelization, we further demonstrate the application of the proposed voxelization in mesh morphing. 

The main contributions of the differentiable voxelization algorithm proposed in this paper are as follows: 
\begin{itemize}
    \item It is a fast and resolution-flexible algorithm for the voxelization of 3D meshes endowed with  GPU acceleration.
    \item It is a differentiable algorithm for the voxelization of 3D meshes, facilitating mesh morphing and other mesh-level optimization and deep learning tasks.
    \item The proposed algorithm achieves state-of-the-art performance on the mesh-to-voxel task in terms of both accuracy and efficiency on the ShapeNet dataset.
\end{itemize}

\section{Differentiable Voxelization}
In this section, we introduce the technical details of the proposed differentiable voxelization algorithm for 3D meshes based on the winding number and solid angles.

\subsection{Generalized Winding Number}
The winding number is a fundamental concept in topology, which is used to determine the number of times a curve winds around a point. The winding number of a curve $\gamma$ around a point $p$ is defined as the number of times the curve winds around the point in the counterclockwise direction. The winding number is a topological invariant, which is invariant under continuous deformation of the curve. The winding number is defined as follows:
\begin{equation}
    W(p, \gamma) = \frac{1}{2\pi} \oint_{\gamma} \frac{\partial \gamma}{\partial t} \cdot \frac{\partial^2 \gamma}{\partial t^2} dt,
\end{equation}

For 3D cases, the winding number of a surface $\Sigma$ around a point $p$ is defined as the number of times the surface winds around the point in the counterclockwise direction. The winding number is a topological invariant, which is invariant under continuous deformation of the surface. The winding number is defined as follows:
\begin{equation}
  \begin{aligned}
    W(q, \Sigma) & = \frac{1}{4\pi}\iint_{\Sigma} \left \langle \vec{n},\vec{E_q} \right \rangle ds, \\
     \vec{E_q}(x) & = \frac{\vec{x} - \vec{q}}{\left \| \vec{x} - \vec{q} \right \|^3}.
  \end{aligned}
\end{equation}
For a closed boundary surface $\Sigma = \partial \Omega$, the winding number around the query point $p$ is equivalent to the occupancy, which is a direct consequence of the Gaussian integral theorem \cite{maxwell1954electricity}, i.e., 
\begin{equation}
  \begin{aligned}
    W(q, \Sigma) = & \frac{1}{4\pi}\iint_{\Sigma} \left \langle \vec{n},\vec{E_q} \right \rangle ds = \frac{1}{4\pi}\iiint_{\Omega} \nabla \cdot \vec{E_q} dV \\
    = & \iiint_{\Omega} \delta(\vec{r} - \vec{q}) dV = \text{Occp}(q, \Sigma),
  \end{aligned}
\end{equation}
where $\delta(\vec{r} - \vec{p})$ is the Dirac delta function. The naive discretization of the winding number for a mesh $(V, F)$ can be formulated as follows:
 can be formulated as follows:
\begin{equation}
  \label{eq:integral}
  \begin{aligned}
    W(q, \Sigma) = & \frac{1}{4\pi}\sum_{i=1}^{F} \left \langle \vec{n}_i,\vec{E_q}(c_i) \right \rangle \Delta_i, \\
  \end{aligned}
\end{equation}
where $n_i$, $c_i$, and $\Delta_i$ are the normal vector, the center, and the area of the $i$-th face, respectively. When the mesh is well-regularized (e.g., triangular mesh constructed by the regular triangles) and the query point is far from the surface, the naive discretization of the winding number is accurate enough. 

\subsection{Solid Angles}
However, general 3D meshes, especially the artifact meshes from manually designing and manufacturing, usually contain sharp edges and irregular triangles, which leads to considerable errors in the naive discretization of the winding number. To address this issue, \cite{jacobson2013robust} proposed the hierarchical winding number evaluation by calculating solid angles, i.e.,
\begin{equation}
  \label{eq:occu}
  \begin{aligned}
    \text{Occp}(q, \Sigma) = W(p, \Sigma) = & \frac{1}{4\pi}\sum_{i=1}^{F} \Omega_i(q),
  \end{aligned}
\end{equation}
where $\Omega_i(q)$ is the solid angle of the $i$-th face at the query point $q$. The solid angle \cite{van1983solid} is defined as the ratio of the area of the spherical cap to the square of the radius of the sphere, i.e.,
\begin{equation}
  \begin{aligned}
    \Omega_i(q) = & \frac{1}{r^2} \iint_{\Omega_i(q)} \cos^{-1} \left \langle \vec{n}_i, \vec{E_q} \right \rangle d\Omega.
  \end{aligned}
\end{equation}
The mathematical equivalence between the solid angle and the winding number can be found in \cite{werner2017solid}. For a triangle with vertices $v_j$, $j = 0, 1, 2$, the solid angle can be computed as follows:
\begin{equation}
  \begin{aligned}
    \Omega(q) & = \arctan\frac{\det(\vec{e}_0, \vec{e}_1, \vec{e}_2)}{1+\left \langle \vec{e}_0, \vec{e}_1 \right \rangle + \left \langle \vec{e}_1, \vec{e}_2 \right \rangle + \left \langle \vec{e}_2, \vec{e}_0 \right \rangle}, \\
    \vec{e}_j &= \frac{\vec{v}_j - \vec{q}}{\left \| \vec{v}_j - \vec{q} \right \|}.
  \end{aligned}
\end{equation}
Then the occupancy of a query point $q$ can be computed as the sum of the solid angles of all the faces from the mesh i.e., Eq. \eqref{eq:occu}.

\begin{figure}[!htp]
  \centering
  \includegraphics[width=0.6\linewidth]{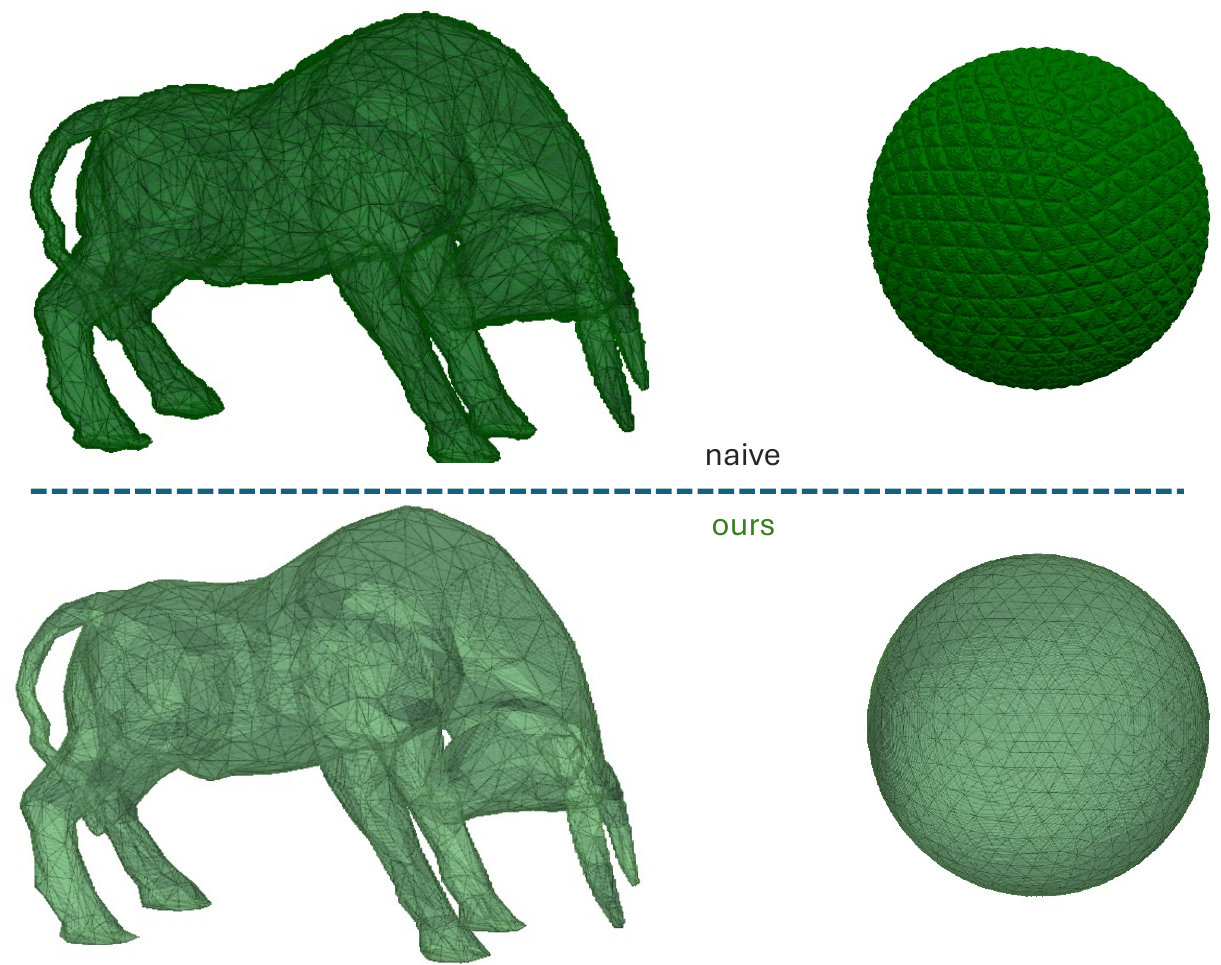}
  \caption{The naive computation of the solid angles like Eq. \eqref{eq:occu} (TOP) leads tufted covers near the original surfaces. The proposed differentiable voxelization algorithm replaces $arctan$ by ${\rm ATAN2}$ to address this issue, which obtains almost binary occupancy and reconstruction in high accuracy (BLOW).}
  \label{fig:artan}
\end{figure}

The naive computation of the solid angles like Eq. \eqref{eq:occu} leads a computational gap near surface boundaries due to the zero point of the denominator, which forms a  tufted cover, shown in Fig. \ref{fig:artan}. \cite{jacobson2013robust} involves discretization and smoothing to achieve binary occupancy, which declines the accuracy. We replace $arctan$ by ${\rm ATAN2}$ \cite{organick1966some} to address this issue, where ${\rm ATAN2}$ is defined as follows:
\begin{equation}
  \begin{aligned}
    {\rm ATAN2}(y, x) = & \begin{cases}
      \arctan\frac{y}{x}, & x > 0, \\
      \arctan\frac{y}{x} + \pi, & x < 0, y \geq 0, \\
      \arctan\frac{y}{x} - \pi, & x < 0, y < 0, \\
      \frac{\pi}{2}, & x = 0, y > 0, \\
      -\frac{\pi}{2}, & x = 0, y < 0, \\
      0, & x = 0, y = 0.
    \end{cases}
  \end{aligned}
\end{equation}
We find
that the ${\rm ATAN2}$ facilitates the occupancy to be almost binary and gradients-vanishing ($0$ or $1$), except the query points on the surface. 
The proposed differentiable voxelization algorithm is presented by Algorithm \ref{alg:voxelization}.
\begin{algorithm}[!htp]
  \caption{Differentiable Voxelization}
  \label{alg:voxelization}
  \begin{algorithmic}
    \REQUIRE Mesh $(V, F)$, query point $q$
    \ENSURE Occupancy ${\rm Occp}(q)$

    Initialize ${\rm Occp}(q) = 0$
    \FOR{each triangle $f_i$ in $F$}
    \STATE $f_i = (v_0, v_1, v_2)$, normalize $e_i = \frac{v_i - q}{\left \| v_i - q \right \|}$

    \STATE $\alpha = \det(e_0, e_1, e_2), \beta = 1+\left \langle e_0, e_1 \right \rangle + \left \langle e_1, e_2 \right \rangle + \left \langle e_2, e_0 \right \rangle$

    \STATE ${\rm Occp}(q) = {\rm Occp}(q) +
    \arctan 2\left(\alpha, \beta\right)$
    \ENDFOR
    \STATE Return ${\rm Occp}(q)$
  \end{algorithmic}
\end{algorithm}

Fig. \ref{fig:diffoccupancy} shows an illustration of the differentiable voxelization of 3D meshes, including non-watertight meshes (almost-closed meshes), meshes with non-regular triangles, and combinational and complex geometry, which causes the failure of traditional voxelization methods. More importantly, the proposed differentiable voxelization algorithm can be easily implemented on GPU, which achieves fast and accurate voxelization of 3D meshes.

\subsection{Voxelization of Open Surfaces}
Although the differentiable voxelization has no limits on non-watertight meshes by automatically filling the miss faces through continuous integration, as shown in Fig. \ref{fig:diffoccupancy} (a), it is still a challenge to voxelize completely open meshes, which is a common case for artist-designed models. To address this issue, we propose the flipped duplication to convert the open meshes into almost closed meshes, described as the following algorithm \ref{alg:duplication}.
\begin{algorithm}[!htp]
  \caption{Flipping Duplication}
  \label{alg:duplication}
  \begin{algorithmic}
    \REQUIRE Open Mesh $\Sigma = (V, F)$
    \ENSURE Flipped Duplication $\Sigma^{-2} =  (V \cup V', F \cup F')$

    \STATE Compute normal vectors $N_i$ for each vertices $v_i$ in $V$
    \STATE Copy the vertices and perturb them towards the negative normal direction, $v' = v_i - \epsilon N_i$, $V' = \{v' | v_i \in V\}$ (where $\epsilon$ is a small positive number, e.g., $0.01$)
    \STATE Copy and flip the faces  \STATE $F' = \{(v_0', v_1', v_2') | f = (v_0, v_1, v_2) \in F\}$

    \STATE Return $\Sigma^{-2} =  (V \cup V', F \cup F')$
  \end{algorithmic}
\end{algorithm}

\begin{figure}[!htp]
  \centering
  \includegraphics[width=\linewidth]{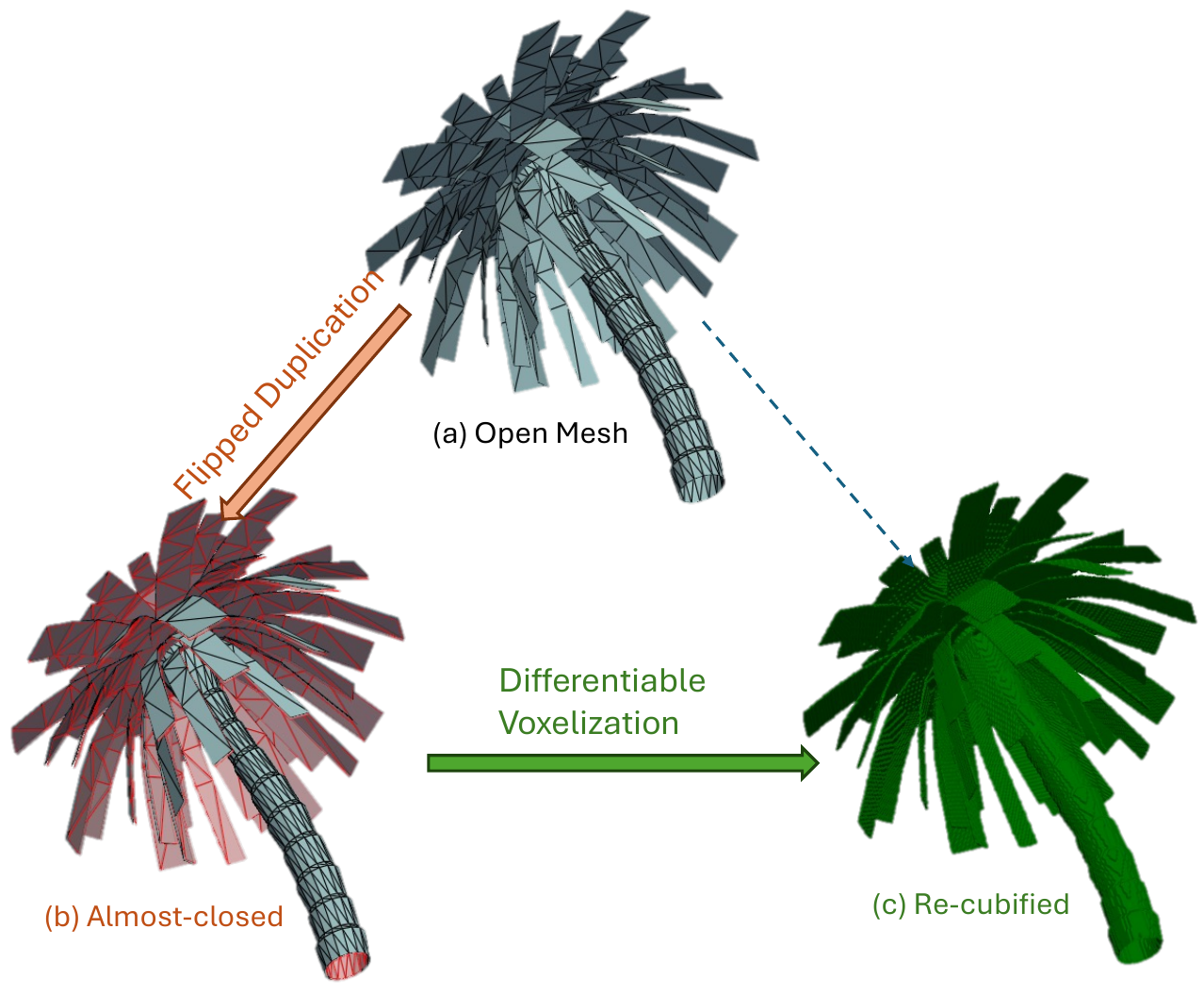}
  \caption{The flipped duplication converts the open meshes into almost closed meshes, which can be voxelized by the proposed differentiable voxelization algorithm.}
  \label{fig:open}
\end{figure}

Apply Algorithm \ref{alg:voxelization} on the flipped duplication $\Sigma^{-2}$ obtained by Algorithm \ref{alg:duplication}, the open meshes can be voxelized as the closed meshes, which is shown in Fig. \ref{fig:open}. The thickness of the slit between the flipped duplication is determined by the perturbation $\epsilon$ in Algorithm \ref{alg:duplication}. Algorithm \ref{alg:voxelization} can easily overcome the non-watertight issue caused by the slit.

\subsection{Mesh Morphing}
\begin{figure*}[!htp]
  \centering
  \includegraphics[width=\linewidth]{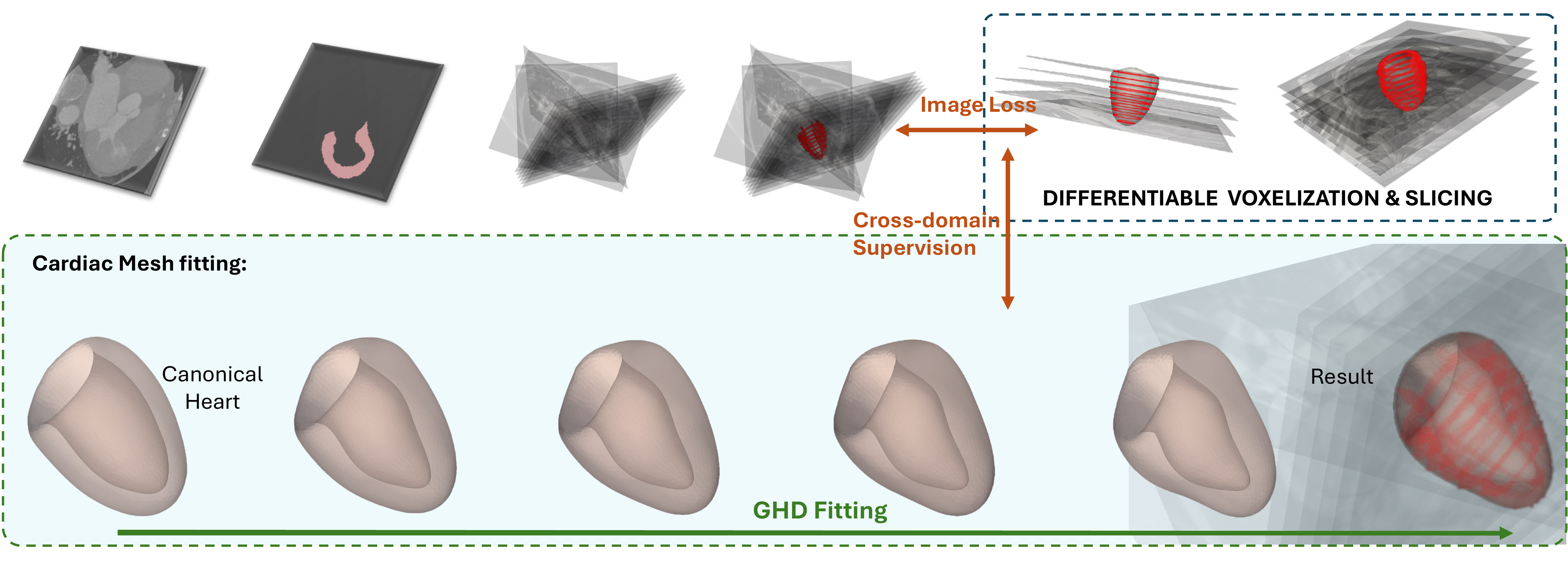}
  \caption{Mesh morphing is a process of deforming a 3D mesh to another 3D mesh by a neural network. The proposed differentiable voxelization facilitates the optimization of mesh morphing, which admits the computation of gradients with respect to the input mesh under the supervision or conditions of voxelization ground truth.}
  \label{fig:morphing}
\end{figure*}

Differentiable voxelization facilitates the optimization of mesh morphing, which admits the computation of gradients with respect to the input mesh under the supervision or conditions of voxelization ground truth. The voxelization ground truth can be usually obtained by CT scans \cite{buzug2011computed}. Therefore the differentiable voxelization lays the foundation for 3D mesh reconstruction from 2D medical scan images \cite{litjens2017survey}. It is worth noting that the differentiable voxelization is not limited to mesh morphing but can be incorporated in other mesh-level optimization and deep learning tasks, such as mesh segmentation, mesh classification, and mesh generation. 

Algorithm \ref{alg:voxelization} provided an accurate differentiable voxelization algorithm for mesh morphing, which means the Eq. \eqref{eq:occu} converges to the ground truth of the voxelization. Even though Eq. \eqref{eq:occu} is composed by differentiable functions, the results approach the binary values of the voxelization ground truth. However, pros always come with cons. High accuracy of binary voxelization leads to gradients vanishing, despite the calculation being differentiable. In the practice of mesh reconstruction of medical images, we use \eqref{eq:integral} to avoid the vanishing gradients, which is a trade-off between the accuracy and the differentiability. Fig \ref{fig:morphing} shows an illustration of the mesh morphing to fit the target segmentation of a human heart. Refer to \cite{luo2023ghd} for more details of the cardiac mesh reconstruction via the differentiable voxelization.

\section{Experiments}

In the experiments, we evaluate our differentiable voxelization algorithm on the Mesh-to-occupancy task compared with other existing methods on selected meshes from \cite{DBLP:conf/siggraph/KrishnamurthyL96} and \cite{Common-3d-test-Models}. The methods evaluated include the "contains" method in Trimesh \cite{trimesh} library, the Mesh-to-SDF library \cite{mesh-to-sdf}, and the SDF computing method used in \cite{Wang_2022DOGN}, where we denoted as DOGN. 

Direct evaluation of the occupancy field is not available due to the lack of ground truth. In fact, a more accurate occupancy field will yield a better reconstruction mesh via marching cubes. We measure the geometric distance between the reconstructed and original mesh to evaluate the mesh-to-voxel algorithms. We use Hausofroff and Chamfer distances as the main metrics. 

We evaluate the methods at different resolutions, from 32 to 256 (32, 64, 128, 256).
The Mesh-to-SDF method has two kinds of surface point methods namely 'scan' and 'sample'. The 'Scan' method, DOGN and our method are evaluated under all resolutions from 32 to 256. The 'Sample' method and 'contains' method are only evaluated under 32 and 64 which we think are sufficient to demonstrate their performance in terms of quality and time.

\subsection{Settings and Details of the Methods}

\begin{itemize}
    \item [1.] Ours: since our method relies on GPU to process the data in parallel, we set the batch size of our method, i.e., the number of query points of each batch, to 2000 due to the memory limitation. Also, all the calculations are done under half-precision.
    \item [2.] Mesh-to-sdf (scan): For 'scan', we used the 'mesh-to-voxels' API and set different resolutions directly.
    \item [3.] Mesh-to-sdf (sample): For the 'sample' method, we set the batch size to 10000.
    \item [4.] Trimesh.contains: For trimesh's contains method, we set the batch size to 10000.
    \item [5.] DOGN: No specific setting for DOGN, the level is set to 2/resolution as recommended.
\end{itemize}

All the meshes are normalized to [-1,1] before the computation of the occupancy filed (SDF). All occupancy fields are firstly cubified and then smoothed with filter Laplacian method in Trimesh with ten iterations and $\lambda = 0.15$. Then, we compute the Chamfer distance and the Hausdorff distance with 20000 samples between the normalized original mesh and smoothed mesh 3 times, with the standard deviation reported. Results are computed under a single Nvidia RTX A6000 GPU and an AMD EPYC 7443P 24-core Processor.

\subsection{Results}

\begin{table*}[h]
\centering
\begin{tabularx}{.85\textwidth}{|>{\raggedright\arraybackslash}p{2cm}|X|X|X|X||X|X|X||X|X|X|}
\hline
Name (Watertight) & Num Faces & \tabularxmulticolumncentered{3}{X||}{\vspace{0pt} \large Chamfer Distance} & \tabularxmulticolumncentered{3}{X||}{\vspace{0pt}  \large Hausdorff Distance} & \tabularxmulticolumncentered{3}{X|}{\vspace{0pt}  \large Time(s)} \\ \cline{3-11} 
 &  & Diff  & 'Scan'  & DOGN  & Diff & 'Scan' & DOGN & Diff & 'Scan' & DOGN \\
\hline
Rocker-arm($\checkmark$) & 20088 & \textbf{0.01702} & 0.05966 & 0.02648 & \textbf{0.02691} & 0.10362 & 0.03337 & 198.46 & 306.67 & 103.14 \\
Beast & 64618 & \textbf{0.01502} & 0.06272 & 0.02526 & \textbf{0.02309} & 0.10318 & 0.08098 & 744.10 & 368.34 & 107.00 \\
Max Planck & 99991 & \textbf{0.02033} & 0.37075 & 0.02555 & 0.18064 & 1.08849 & \textbf{0.03709} & 1158.30 & 484.46 & 135.59 \\
Lucy($\checkmark$) & 99970 & \textbf{0.01501} & 0.04233 & 0.02463 & \textbf{0.02503} & 0.11022 & 0.04462 & 1155.86 & 345.91 & 111.01 \\
Cheburashka & 13334 & \textbf{0.01806} & 0.06823 & 0.02769 & \textbf{0.02977} & 0.16839 & 0.08520 & 130.68 & 392.88 & 121.58 \\
Happy & 98601 & \textbf{0.01726} & 0.06299 & 0.02656 & \textbf{0.02687} & 0.15838 & 0.03588 & 1140.47 & 319.96 & 114.05 \\
Teapot & 6320 & \textbf{0.01757} & 0.25775 & 0.02353 & 0.10390 & 1.01302 & \textbf{0.03257} & 61.94 & 462.17 & 114.65 \\
Bimba($\checkmark$) & 224906 & \textbf{0.02056} & 0.09320 & 0.02989 & \textbf{0.03400} & 0.11556 & 0.03926 & 3050.98 & 455.75 & 135.22 \\
Igea($\checkmark$) & 268686 & \textbf{0.02164} & 0.13443 & 0.03212 & 0.04057 & 0.10788 & \textbf{0.03991} & 3649.88 & 534.83 & 135.75 \\
Hourse($\checkmark$) & 96966 & \textbf{0.01587} & 0.08751 & 0.02600 & \textbf{0.02425} & 0.60148 & 0.03771 & 1120.32 & 418.90 & 115.71 \\
Homer($\checkmark$) & 12000 & \textbf{0.01507} & 0.05253 & 0.02510 & \textbf{0.02412} & 0.09813 & 0.07579 & 117.49 & 341.66 & 107.74 \\
Spot & 5856 & \textbf{0.01996} & 0.08974 & 0.02957 & \textbf{0.03278} & 0.11833 & 0.03788 & 57.42 & 441.26 & 120.16 \\
Armadillo($\checkmark$) & 99976 & \textbf{0.01876} & 0.07367 & 0.02858 & \textbf{0.03146} & 0.11894 & 0.03571 & 1155.40 & 356.79 & 121.72 \\
Suzanne & 968 & \textbf{0.01947} & 0.09038 & 0.03012 & \textbf{0.08047} & 0.58337 & 0.09545 & 9.79 & 430.40 & 108.23 \\
Stanford Bunny & 69451 & \textbf{0.02215} & 0.20425 & 0.02847 & 0.10349 & 0.84211 & \textbf{0.04049} & 802.02 & 479.98 & 130.56 \\
Nefertiti($\checkmark$) & 99938 & \textbf{0.01779} & 0.08017 & 0.02747 & \textbf{0.03064} & 0.10384 & 0.03866 & 1156.84 & 560.41 & 118.79 \\
Ogre & 124008 & \textbf{0.01711} & 0.06440 & 0.02906 & \textbf{0.07934} & 0.13050 & 0.12526 & 1441.33 & 375.40 & 112.09 \\
Cow($\checkmark$) & 5804 & \textbf{0.01534} & 0.21486 & 0.02528 & \textbf{0.02932} & 0.95404 & 0.04360 & 57.02 & 354.76 & 105.84 \\
Fandisk($\checkmark$) & 12946 & \textbf{0.02054} & 0.08235 & 0.03002 & \textbf{0.03789} & 0.48765 & 0.04194 & 127.07 & 321.94 & 107.73 \\
Car & 157384 & \textbf{0.03552} & 0.07919 & 0.04584 & 0.16396 & \textbf{0.16026} & 0.18403 & 1849.66 & 285.01 & 129.18 \\
Tableware & 4788 & \textbf{0.01938} & 0.05631 & 0.20433 & \textbf{0.04357} & 0.17771 & 0.70841 & 47.18 & 247.94 & 102.09 \\
Ladder & 828 & \textbf{0.01530} & 0.06477 & 0.02357 & \textbf{0.02224} & 0.12011 & 0.03213 & 16.29 & 156.47 & 101.86 \\
Bull & 2418 & \textbf{0.01802} & 0.21368 & 0.02774 & \textbf{0.02930} & 0.81471 & 0.03632 & 24.04 & 373.57 & 112.22 \\
\hline
\end{tabularx}
\caption{Results and Computation Times of Different Methods for resolution 256}
\label{table_resol256nostd}
\end{table*}

It can be seen from table \ref{table_resol256nostd} that our method significantly outperforms 'Scan' and DOGN in terms of quality (for results with standard deviation, please refer to table \ref{table_resol256} in Appendix \ref{appendix}), under the resolution of 256. In cases where the two distances have a large difference, 'Car' for instance, is because the original mesh is non-watertight, and our method fill those holes and splits. The resulting mesh still captures the details of the original mesh and looks similar,  shown in Fig. \ref{fig:diffoccupancy} (c). The time cost of our method is closely related to the number of faces of the mesh and the memory of the GPU, since our method relies on GPU memory, and computes the integral over all faces of a mesh. For example, our method processes mesh 'Ladder' and 'Suzanne' faster than the other two methods, but is significantly slower then process complex meshes like 'Igea'. However, with sufficient computing resources, our method will be significantly faster as shown in table \ref{resol32}. Otherwise, the processing time grows almost linearly to the number of faces and cubicly to the resolution when computing resources are limited.

For lower resolutions, we also include the 'contains' and 'sample' methods into consideration, our method produces comparable results to other methods as shown in table \ref{resol32} and \ref{resol64}, \ref{resol32_64}.

\bibliographystyle{ACM-Reference-Format}
\bibliography{ref}

\newpage
\appendix
\section{Appendix}\label{appendix}

\begin{table*}[h]
\centering
\begin{tabularx}{\textwidth}{|>{\raggedright\arraybackslash}p{2cm}|X|X|X|X|X|X||X|X|X|X|X|}
\hline
Name (watertight)& Num Faces & \tabularxmulticolumncentered{5}{X||}{\vspace{0pt} \large Chamfer Distance(std)} & \tabularxmulticolumncentered{5}{X|}{\vspace{0pt}  \large Time(s)} \\ \cline{3-12} 
 &  & Diff  & 'Scan'  & Trimesh  & 'Sample'  & DOGN  &  Diff &  'Scan' &  Trimesh& 'Sample' &  DOGN \\
\hline
Rocker-arm($\checkmark$) & 20088 & \textbf{0.06632} (0.00014) & 0.08379 (0.00013) & 0.06656 (0.00013) & 0.06641 (0.00008) & 0.15448 (0.00035) & 0.4251 & 12.3401 & 10.3270 & 72.9459 & 0.3631 \\
Beast & 64618 & \textbf{0.07167} (0.00011) & 0.10516 (0.00032) & 0.07189 (0.00019) & 0.07183 (0.00034) & 0.15194 (0.00068) & 0.0075 & 14.3608 & 18.5567 & 72.6238 & 0.4087 \\
Max Planck & 99991 & \textbf{0.06057} (0.00013) & 0.35081 (0.00080) & 0.06569 (0.00022) & 0.06266 (0.00009) & 0.11371 (0.00066) & 0.0065 & 16.9182 & 102.0209 & 76.7355 & 0.5564 \\
Lucy($\checkmark$) & 99970 & 0.05977 (0.00016) & 0.07760 (0.00038) & \textbf{0.05945} (0.00004) & 0.05951 (0.00026) & 0.15320 (0.00027) & 0.0062 & 16.3613 & 66.0559 & 71.7929 & 0.4569 \\
Cheburashka($\checkmark$) & 13334 & \textbf{0.06254} (0.00012) & 0.08655 (0.00028) & 0.06275 (0.00019) & 0.06270 (0.00024) & 0.16430 (0.00055) & 0.0045 & 12.6865 & 3.1635 & 72.2148 & 0.3245 \\
Happy & 98601 & 0.05548 (0.00013) & 0.08143 (0.00032) & 0.05548 (0.00021) & \textbf{0.05537} (0.00007) & 0.16028 (0.00066) & 0.0073 & 17.0295 & 23.5146 & 71.7547 & 0.4803 \\

Teapot & 6320 & \textbf{0.05685} (0.00006) & 0.12914 (0.00071) & 0.05716 (0.00021) & 0.06477 (0.00012) & 0.16999 (0.00031) & 0.0046 & 11.7145 & 4.4557 & 72.3538 & 0.3307 \\

Bimba($\checkmark$) & 224906 & 0.05728 (0.00014) & 0.10217 (0.00025) & \textbf{0.05718} (0.00021) & \textbf{0.05718} (0.00013) & 0.16595 (0.00026) & 0.3337 & 20.9800 & 244.5197 & 75.5195 & 0.7512 \\

Igea($\checkmark$) & 268686 & 0.06211 (0.00006) & 0.13745 (0.00031) & \textbf{0.06203} (0.00016) & 0.06215 (0.00003) & 0.18204 (0.00030) & 3.6248 & 24.0900 & 469.7210 & 78.1245 & 0.8369 \\

Hourse($\checkmark$) & 96966 & \textbf{0.06970 (0.00013)} & 0.12735 (0.00030) & 0.07019 (0.00007) & 0.10821 (0.00067) & 0.15961 (0.00018) & 0.0072 & 16.3146 & 87.9119 & 74.2368 & 0.4593 \\

Homer($\checkmark$) & 12000 & 0.05890 (0.00005) & 0.07347 (0.00025) & 0.05876 (0.00017) & \textbf{0.05841} (0.00020) & 0.15538 (0.00039) & 0.0063 & 11.2691 & 2.3533 & 74.2110 & 0.3246 \\

Spot & 5856 & \textbf{0.06440} (0.00007) & 0.10161 (0.00046) & 0.06463 (0.00017) & 0.06449 (0.00016) & 0.16768 (0.00032) & 0.0062 & 11.5103 & 9.6123 & 69.0767 & 0.3217 \\

Armadillo($\checkmark$) & 99976 & 0.07081 (0.00001) & 0.11524 (0.00024) & \textbf{0.07046} (0.00028) & 0.07061 (0.00021) & 0.16688 (0.00056) & 0.0055 & 16.8934 & 121.1818 & 71.5409 & 0.5164 \\

Suzanne & 968 & 0.06440 (0.00024) & 0.10288 (0.00031) & 0.06312 (0.00003) & \textbf{0.06173} (0.00017) & 0.17534 (0.00040) & 0.0064 & 12.1686 & 1.2807 & 69.1196 & 0.3132 \\
Stanford Bunny & 69451 & \textbf{0.06344} (0.00011) & 0.15924 (0.00045) & 0.06814 (0.00024) & 0.06364 (0.00020) & 0.12063 (0.00026) & 0.0074 & 16.2307 & 81.7012 & 71.1485 & 0.5156 \\
Nefertiti($\checkmark$) & 99938 & 0.05991 (0.00011) & 0.08721 (0.00039) & \textbf{0.05972} (0.00022) & 0.06000 (0.00023) & 0.16238 (0.00036) & 0.0055 & 17.6811 & 104.9907 & 75.9086 & 0.4892 \\
Ogre & 124008 & 0.06791 (0.00030) & 0.09158 (0.00013) & \textbf{0.06654} (0.00024) & 0.06862 (0.00033) & 0.15788 (0.00040) & 0.0086 & 18.0386 & 12.8931 & 72.2672 & 0.5144 \\
Cow($\checkmark$) & 5804 & \textbf{0.06242} (0.00011) & 0.21833 (0.00130) & 0.06272 (0.00007) & 0.17479 (0.00264) & 0.15892 (0.00036) & 0.0048 & 12.1376 & 1.2943 & 69.5902 & 0.2743 \\
Fandisk($\checkmark$) & 12946 & \textbf{0.05456} (0.00016) & 0.09516 (0.00026) & 0.05460 (0.00005) & 0.05467 (0.00001) & 0.16631 (0.00063) & 0.0047 & 13.4979 & 6.9413 & 69.3316 & 0.3324 \\
Car& 157384& 0.07015 (0.00014)&  0.08850 (0.00011) &  \textbf{0.06765} (0.00013) & 0.07037 (0.00034)&0.23775 (0.00040)& 3.8466 & 20.1299 &799.0143 & 76.3597 & 0.7136 \\
Tableware&  4788& 0.63220 (0.00228) & 0.21702 (0.00033) &  0.62292 (0.00127)& 0.62329 (0.00116) & \textbf{0.17387} (0.00065) &0.0062 &11.5668 & 3.6265& 72.8735& 0.3134\\
Ladder & 828 & 0.28433 (0.00046) & 0.29627 (0.00090)  &0.28649 (0.00112) & 0.28678 (0.00136) & \textbf{0.13036} (0.00032) & 0.1531 &12.5374 & 0.08043 & 70.7119 & 0.2595\\
Bull & 2418 & 0.06685 (0.00029) & 0.21340 (0.00072) & \textbf{0.06670} (0.00023) & 0.13797 (0.00081)&0.16155 (0.00028) &  0.0047& 11.6457& 2.0163&74.1523 &0.3395\\
\hline
\end{tabularx}
\caption{Resulting Chamfer Distances and Computation Times of Different Methods for resolution 32}
\label{resol32}
\end{table*}

\begin{table*}[!h]
\centering
\begin{tabularx}{\textwidth}{|>{\raggedright\arraybackslash}p{2cm}|X|X|X|X|X|X||X|X|X|X|X|}
\hline
Name (watertight) & Num Faces & \tabularxmulticolumncentered{5}{X||}{\vspace{0pt} \large Chamfer Distance(std)} & \tabularxmulticolumncentered{5}{X|}{\vspace{0pt}  \large Time(s)} \\ \cline{3-12} 
 &  & Diff  & 'Scan'  & Trimesh  & 'Sample'  & DOGN  &  Diff &  'Scan' &  Trimesh& 'Sample' &  DOGN \\
\hline
Rocker-arm($\checkmark$) & 20088 & \textbf{0.03568} (0.00009) & 0.06522 (0.00018) & 0.03569 (0.00005) & 0.03575 (0.00010) & 0.08130 (0.00028) & 3.10 & 15.80 & 97.47 & 470.83 & 1.90 \\
Beast & 64618 & 0.03357 (0.00002) & 0.07294 (0.00014) & \textbf {0.03351} (0.00006) & 0.03761 (0.00013) & 0.07956 (0.00020) & 9.04 & 18.61 & 147.10 & 479.61 & 2.04 \\
Max Planck & 99991 & \textbf{0.03495} (0.00006) & 0.35516 (0.00073) & 0.03995 (0.00022) & 0.03806 (0.00009) & 0.05998 (0.00022) & 14.30 & 22.87 & 806.17 & 500.46 & 2.56 \\
Lucy($\checkmark$) & 99970 & \textbf{0.03114} (0.00005) & 0.05217 (0.00014) & 0.03122 (0.00006) & 0.03122 (0.00011) & 0.07905 (0.00018) & 14.07 & 19.51 & 574.55 & 479.88 & 1.99 \\
Cheburashka($\checkmark$) & 13334 & 0.03594 (0.00007) & 0.07236 (0.00033) & \textbf{0.03584} (0.00005) & 0.03607 (0.00010) & 0.08458 (0.00022) & 1.46 & 16.67 & 26.07 & 474.34 & 2.16 \\
Happy & 98601 & 0.03270 (0.00014) & 0.06755 (0.00029) & \textbf{0.03266} (0.00004) & 0.03296 (0.00007) & 0.07941 (0.00042) & 13.89 & 20.64 & 188.29 & 488.89 & 2.17 \\
Teapot & 6320 & \textbf{0.03291} (0.00014) & 0.16490 (0.00093) & 0.03879 (0.00005) & 0.04875 (0.00032) & 0.08601 (0.00013) & 0.70 & 19.10 & 35.85 & 480.15 & 2.01 \\
Bimba($\checkmark$) & 224906 & \textbf{0.03427} (0.00007) & 0.09563 (0.00037) & 0.03432 (0.00007) & 0.03445 (0.00012) & 0.08473 (0.00016) & 41.03 & 27.55 & 2199.34 & 497.14 & 2.74 \\
Igea($\checkmark$) & 268686 & \textbf{0.03663} (0.00006) & 0.13448 (0.00028) & * & 0.03677 (0.00004) & 0.09399 (0.00016) & 52.59 & 31.82 & $>$4000 & 507.74 & 2.86 \\
Hourse($\checkmark$) & 96966 & \textbf{0.03598} (0.00003) & 0.09547 (0.00043) & 0.03605 (0.00011) & 0.06874 (0.00072) & 0.08316 (0.00024) & 15.53 & 22.91 & 791.05 & 484.71 & 2.14 \\
Homer($\checkmark$) & 12000 & 0.03298 (0.00012) & 0.05691 (0.00013) & \textbf{0.03292} (0.00010) & \textbf{0.03292} (0.00007) & 0.07907 (0.00023) & 1.32 & 15.76 & 16.25 & 472.65 & 1.89 \\
Spot & 5856 & 0.03628 (0.00006) & 0.09217 (0.00013) & \textbf{0.03627} (0.00009) & 0.03637 (0.00007) & 0.08544 (0.00015) & 0.65 & 19.74 & 81.19 & 478.09 & 2.05 \\
Armadillo($\checkmark$) & 99976 & 0.03737 (0.00005) & 0.08675 (0.00014) & \textbf{0.03718} (0.00003) & 0.03747 (0.00005) & 0.08546 (0.00012) & 14.07 & 22.15 & 1001.82 & 478.16 & 2.32 \\
Suzanne & 968 & \textbf{0.03614} (0.00006) & 0.09256 (0.00036) & 0.03983 (0.00010) & 0.03779 (0.00004) & 0.08982 (0.00033) & 0.11 & 17.65 & 11.48 & 474.98 & 1.87 \\
Stanford Bunny & 69451 & \textbf{0.03847} (0.00014) & 0.16958 (0.00151) & 0.04645 (0.00009) & 0.03997 (0.00008) & 0.06325 (0.00006) & 9.76 & 22.77 & 669.95 & 475.62 & 2.50 \\
Nefertiti($\checkmark$) & 99938 & 0.03345 (0.00007) & 0.07886 (0.00014) & \textbf{0.03337} (0.00005) & 0.03340 (0.00004) & 0.08305 (0.00033) & 14.07 & 24.16 & 908.54 & 503.51 & 2.30 \\
Ogre & 124008 & 0.03485 (0.00010) & 0.07289 (0.00002) & \textbf{0.03426} (0.00002) & 0.03705 (0.00017) & 0.08312 (0.00016) & 19.13 & 21.47 & 126.22 & 481.03 & 2.20 \\
Cow($\checkmark$) & 5804 & \textbf{0.03376} (0.00008) & 0.20966 (0.00108) & 0.03378 (0.00014) & 0.20101 (0.00081) & 0.07951 (0.00016) & 0.64 & 16.63 & 8.24 & 462.41 & 1.82 \\
Fandisk($\checkmark$) & 12946 & 0.04016 (0.00002) & 0.08417 (0.00022) & \textbf{0.04015} (0.00004) & 0.04017 (0.00004) & 0.07795 (0.00006) & 1.42 & 16.30 & 61.16 & 462.34 & 1.86 \\
Car & 157384&  0.04826 (0.00005)& 0.08326 (0.00029) & \textbf{0.04268} (0.00011)& 0.05191 (0.00005) & 0.14442 (0.00060) & 28.38&  23.19 & 2187.78 & 518.36 & 2.89\\
Tableware &  4788& \textbf{0.11128} (0.00041) & 0.18683 (0.00079) & 0.11143 (0.00022)& 0.11268 (0.00042)& 0.11983 (0.00033)&  0.53 &15.02 & 36.45 & 496.58&1.79\\
 Ladder &  828& 0.13856 (0.00010) & 0.23965 (0.00057)  & 0.13874 (0.00061)& 0.13813 (0.00036) &\textbf{0.07570} (0.00018)   &0.39 & 14.08 &  0.25 &482.66 &1.60\\
 Bull&  2418&0.03542 (0.00003) & 0.19739 (0.00125) &  \textbf{0.03541} (0.00003) &0.15641 (0.00064)&0.08279 (0.00019) &0.27 & 16.94 & 18.02& 521.10& 2.13\\
\hline
\end{tabularx}
\caption{Resulting Chamfer Distances and Computation Times of Different Methods for resolution 64}
\label{resol64}
\end{table*}

\begin{table*}[!h]
\centering
\begin{tabularx}{\textwidth}{|>{\raggedright\arraybackslash}p{2cm}|X|X|X|X|X||X|X|X|X|X|X|}
\hline
Name (watertight) & \tabularxmulticolumncentered{5}{X||}{\vspace{0pt} \large Hausdorff Distance (resol 32)} & \tabularxmulticolumncentered{5}{X|}{\vspace{0pt} \large Hausdorff Distance (resol 64)} \\ \cline{1-11} 
 & Diff  & 'Scan'  & 'Sample'  & 'Trimesh'  & DOGN & Diff  & 'Scan'  & 'Sample'  & 'Trimesh'  & DOGN \\
\hline
Rocker-arm&0.09804 (0.00028)&0.13694 (0.00051)&0.09833 (0.00058)&\textbf{0.09783} (0.00067)&0.21022 (0.00181)&0.04585 (0.00121)&0.11573 (0.00410)&\textbf{0.04443} (0.00119)&0.04651 (0.00053)&0.11563 (0.00050)\\
Beast&0.19894 (0.00107)&0.26045 (0.00093)&\textbf{0.19595} (0.00437)&0.20010 (0.00071)&0.23835 (0.00123)&0.08119 (0.00133)&0.13537 (0.00088)&\textbf{0.07804} (0.00296)&0.45707 (0.00260)&0.11417 (0.00135)\\
Max Planck&\textbf{0.17962} (0.00315)&1.09022 (0.00811)&0.27223 (0.00228)&0.22204 (0.00199)&0.21338 (0.00105)&0.17761 (0.00241)&1.07249 (0.00551)&0.26696 (0.00174)&0.63863 (0.00107)&\textbf{0.10693} (0.00124)\\
Lucy&\textbf{0.15998} (0.00132)&0.16109 (0.00193)&0.16288 (0.00137)&0.16278 (0.00154)&0.21806 (0.00112)&\textbf{0.06911} (0.00104)&0.12530 (0.00233)&0.07138 (0.00029)&0.07018 (0.00013)&0.12193 (0.00040)\\
Cheburashka&0.11699 (0.00068)&0.15871 (0.00131)&\textbf{0.11667} (0.00124)&0.11675 (0.00073)&0.22699 (0.00259)&\textbf{0.08833} (0.00058)&0.13572 (0.00356)&0.08870 (0.00190)&0.17981 (0.00110)&0.14794 (0.00174)\\
Happy&0.12999 (0.00068)&0.14404 (0.00079)&0.12847 (0.00076)&\textbf{0.12702} (0.00251)&0.23513 (0.00134)&\textbf{0.06669} (0.00260)&0.12116 (0.00037)&0.07515 (0.00332)&0.13735 (0.00163)&0.11438 (0.00161)\\
Teapot&\textbf{0.11404} (0.00119)&1.03774 (0.00159)&0.29660 (0.00184)&0.29425 (0.00041)&0.27024 (0.00118)&\textbf{0.11140} (0.00194)&1.03101 (0.01343)&0.25777 (0.00379)&0.35074 (0.00100)&0.15234 (0.00113)\\
Bimba&0.10842 (0.00251)&0.17430 (0.00183)&\textbf{0.10810} (0.00163)&0.13392 (0.00679)&0.27758 (0.00387)&0.05352 (0.00076)&0.14291 (0.00019)&\textbf{0.05123} (0.00114)&0.12672 (0.05122)&0.13154 (0.00089)\\
Igea&\textbf{0.08470} (0.00083)&0.16438 (0.00067)&0.08807 (0.00486)&0.08518 (0.00137)&0.22979 (0.00164)&0.04855 (0.00181)&0.12874 (0.00058)&*&\textbf{0.04498} (0.00081)&0.11116 (0.00069)\\
Horse&0.12941 (0.00116)&0.66342 (0.00346)&\textbf{0.12695} (0.00019)&0.69284 (0.00014)&0.22085 (0.00111)&0.05391 (0.00248)&0.62432 (0.00143)&\textbf{0.05171} (0.00058)&0.70203 (0.00186)&0.10906 (0.00110)\\
Homer&0.09541 (0.00109)&0.14960 (0.00024)&0.09477 (0.00110)&\textbf{0.09332} (0.00086)&0.20811 (0.00181)&0.05552 (0.00177)&0.12214 (0.00030)&\textbf{0.05250} (0.00038)&0.05271 (0.00154)&0.13304 (0.00102)\\
Spot&\textbf{0.09370} (0.00125)&0.14076 (0.00066)&0.09496 (0.00147)&0.09580 (0.00093)&0.21575 (0.00071)&0.04495 (0.00060)&0.12255 (0.00149)&\textbf{0.04177} (0.00042)&0.04207 (0.00109)&0.10402 (0.00070)\\
Armadillo&0.16884 (0.00294)&0.18805 (0.00160)&\textbf{0.16510} (0.00644)&0.16510 (0.00147)&0.22296 (0.00237)&0.07307 (0.00485)&0.13891 (0.00158)&\textbf{0.06400} (0.00390)&0.54865 (0.00217)&0.10959 (0.00144)\\
Suzanne&\textbf{0.12759} (0.00192)&0.59997 (0.00491)&0.19467 (0.00321)&0.37024 (0.00129)&0.27440 (0.00019)&\textbf{0.09585} (0.00327)&0.58307 (0.00278)&0.23128 (0.00116)&0.48175 (0.00127)&0.15865 (0.00179)\\
Stanford Bunny&\textbf{0.10395} (0.00281)&0.70338 (0.00362)&0.21473 (0.00037)&0.23383 (0.00432)&0.20713 (0.00074)&\textbf{0.09856} (0.00482)&0.84165 (0.01241)&0.23335 (0.00226)&0.36202 (0.00408)&0.10329 (0.00120)\\
Nefertiti&\textbf{0.09277} (0.00073)&0.14624 (0.00248)&0.09496 (0.00052)&0.09448 (0.00147)&0.21426 (0.00053)&\textbf{0.05853} (0.00330)&0.11916 (0.00113)&0.06024 (0.00078)&0.27661 (0.12798)&0.10759 (0.00192)\\
Ogre&0.15168 (0.00076)&0.16781 (0.00221)&\textbf{0.15102} (0.00196)&0.15623 (0.00155)&0.21357 (0.00012)&0.09467 (0.00219)&0.12800 (0.00029)&\textbf{0.06594} (0.00024)&0.16643 (0.00068)&0.15010 (0.00145)\\
Cow&0.17486 (0.00357)&1.02813 (0.00055)&\textbf{0.17444} (0.00070)&1.05163 (0.00121)&0.21426 (0.00136)&\textbf{0.08899} (0.00295)&0.96566 (0.00043)&0.09079 (0.00102)&0.99838 (0.00161)&0.10803 (0.00122)\\
Fandisk&0.10267 (0.00184)&0.12448 (0.00245)&\textbf{0.09604} (0.00439)&0.10156 (0.00174)&0.20640 (0.00104)&0.06248 (0.00573)&0.22357 (0.00603)&\textbf{0.05985} (0.00411)&0.06077 (0.00293)&0.11238 (0.00552)\\
Car&0.18345 (0.00453)&\textbf{0.17601} (0.00163)&0.18431 (0.00303)&0.23542 (0.00072)&0.50652 (0.00330)&0.17009 (0.00174)&0.16606 (0.00402)&\textbf{0.10175} (0.00323)&0.24027 (0.00020)&0.41532 (0.00032)\\
Tableware&0.89475 (0.00284)&\textbf{0.44641} (0.00864)&0.86789 (0.00376)&0.86469 (0.00213)&0.64450 (0.00150)&\textbf{0.21437} (0.00086)&0.38609 (0.00240)&0.22125 (0.00089)&0.41498 (0.00446)&0.67902 (0.00256)\\
Ladder&0.52241 (0.00009)&0.54227 (0.00024)&0.52273 (0.00013)&0.52240 (0.00060)&\textbf{0.27069} (0.00046)&0.21872 (0.00060)&0.37963 (0.00067)&0.21957 (0.00044)&0.21828 (0.00047)&\textbf{0.13241} (0.00110)\\
Bull&0.13289 (0.00373)&0.80594 (0.00121)&\textbf{0.13217} (0.00162)&0.74641 (0.00350)&0.21890 (0.00068)&0.07061 (0.00363)&0.80143 (0.00205)&\textbf{0.06779} (0.00362)&0.79966 (0.00378)&0.10765 (0.00041)\\

\hline
\end{tabularx}
\caption{Resulting Hausdorff Distances of Different Methods for resolutions 32 and 64}
\label{resol32_64}
\end{table*}

\begin{table*}[h]
\centering
\begin{tabularx}{\textwidth}{|>{\raggedright\arraybackslash}p{2cm}|X|X|X|X||X|X|X||X|X|X|}
\hline
Name (watertight) & Num Faces & \tabularxmulticolumncentered{3}{X||}{\vspace{0pt} \large Chamfer Distance (std)} & \tabularxmulticolumncentered{3}{X||}{\vspace{0pt} \large Hausdorff Distance (std)} & \tabularxmulticolumncentered{3}{X|}{\vspace{0pt} \large Time(s)} \\ \cline{3-11} 
 &  & Diff  & 'Scan'  & DOGN  & Diff & 'Scan' & DOGN & Diff & 'Scan' & DOGN \\
\hline
Rocker-arm($\checkmark$) & 20088 & \textbf{0.02212} (0.00001) & 0.06035 (0.00007) & 0.04405 (0.00012) & \textbf{0.02875} (0.00090) & 0.10700 (0.00228) & 0.05536 (0.00158) & 24.91 & 48.49 & 14.67 \\
Beast & 64618 & \textbf{0.02020} (0.00003) & 0.06433 (0.00011) & 0.04267 (0.00005) & \textbf{0.04024} (0.00282) & 0.11064 (0.00161) & 0.08628 (0.00031) & 89.32 & 58.26 & 15.20 \\
Max Planck & 99991 & \textbf{0.02451} (0.00010) & 0.36514 (0.00189) & 0.03628 (0.00008) & 0.17925 (0.00243) & 1.07719 (0.00331) & \textbf{0.05734} (0.00065) & 139.89 & 75.70 & 18.59 \\
Lucy($\checkmark$) & 99970 & \textbf{0.01961} (0.00005) & 0.04413 (0.00020) & 0.04246 (0.00008) & \textbf{0.03140} (0.00191) & 0.11078 (0.00200) & 0.07159 (0.00016) & 139.63 & 52.06 & 13.55 \\
Cheburashka($\checkmark$) & 13334 & \textbf{0.02283} (0.00002) & 0.06890 (0.00015) & 0.04588 (0.00010) & \textbf{0.03618} (0.00377) & 0.12105 (0.00101) & 0.10863 (0.00202) & 15.71 & 60.38 & 16.87 \\
Happy & 98601 & \textbf{0.02165} (0.00007) & 0.06306 (0.00022) & 0.04383 (0.00008) & \textbf{0.03353} (0.00252) & 0.15536 (0.00016) & 0.05918 (0.00079) & 137.77 & 55.52 & 15.53 \\
Teapot & 6320 & \textbf{0.02224} (0.00002) & 0.21243 (0.00083) & 0.04506 (0.00018) & 0.10466 (0.00050) & 1.00711 (0.00192) & \textbf{0.10269} (0.00028) & 7.47 & 68.04 & 14.69 \\
Bimba($\checkmark$) & 224906 & \textbf{0.02426} (0.00006) & 0.09357 (0.00016) & 0.04720 (0.00009) & \textbf{0.03871} (0.00186) & 0.12011 (0.00015) & 0.06437 (0.00091) & 374.64 & 75.76 & 17.81 \\
Igea($\checkmark$) & 268686 & \textbf{0.02575} (0.00006) & 0.13466 (0.00015) & 0.05147 (0.00011) & \textbf{0.03688} (0.00157) & 0.11490 (0.00136) & 0.06003 (0.00266) & 451.48 & 87.85 & 18.27 \\
Hourse($\checkmark$) & 96966 & \textbf{0.02130} (0.00000) & 0.08847 (0.00019) & 0.04442 (0.00012) & \textbf{0.03175} (0.00265) & 0.60367 (0.00467) & 0.05657 (0.00138) & 135.42 & 66.81 & 14.92 \\
Homer($\checkmark$) & 12000 & \textbf{0.02014} (0.00001) & 0.05297 (0.00021) & 0.04231 (0.00009) & \textbf{0.02928} (0.00239) & 0.10570 (0.00048) & 0.09825 (0.00138) & 14.13 & 53.53 & 13.99 \\
Spot & 5856 & \textbf{0.02442} (0.00010) & 0.09000 (0.00012) & 0.04708 (0.00004) & \textbf{0.03489} (0.00044) & 0.11879 (0.00149) & 0.05470 (0.00110) & 6.93 & 66.23 & 14.88 \\
Armadillo($\checkmark$) & 99976 & \textbf{0.02365} (0.00009) & 0.07687 (0.00006) & 0.04697 (0.00010) & \textbf{0.03633} (0.00140) & 0.12065 (0.00033) & 0.05840 (0.00132) & 139.45 & 61.74 & 15.53 \\
Suzanne & 968 & \textbf{0.02423} (0.00001) & 0.09001 (0.00010) & 0.04917 (0.00018) & \textbf{0.08368} (0.00175) & 0.58039 (0.00234) & 0.11331 (0.00174) & 1.19 & 64.79 & 13.75 \\
Stanford Bunny & 69451 & \textbf{0.02668} (0.00004) & 0.18646 (0.00057) & 0.03927 (0.00010) & 0.09665 (0.00154) & 0.79551 (0.00374) & \textbf{0.06049} (0.00240) & 96.41 & 71.52 & 17.91 \\
Nefertiti($\checkmark$) & 99938 & \textbf{0.02221} (0.00003) & 0.07914 (0.00014) & 0.04485 (0.00004) & \textbf{0.03048} (0.00077) & 0.10623 (0.00101) & 0.06128 (0.00267) & 139.66 & 81.66 & 15.88 \\
Ogre & 124008 & \textbf{0.02241} (0.00004) & 0.06616 (0.00023) & 0.04644 (0.00016) & \textbf{0.09947} (0.00153) & 0.12944 (0.00218) & 0.13369 (0.00087) & 175.89 & 59.81 & 14.76 \\
Cow($\checkmark$) & 5804 & \textbf{0.02030} (0.00002) & 0.21046 (0.00219) & 0.04294 (0.00011) & \textbf{0.03217} (0.00083) & 0.95843 (0.00681) & 0.05663 (0.00025) & 6.88 & 52.27 & 13.42 \\
Fandisk($\checkmark$) & 12946 & \textbf{0.02753} (0.00005) & 0.08065 (0.00017) & 0.04490 (0.00013) & \textbf{0.03944} (0.00156) & 0.47152 (0.00420) & 0.06074 (0.00087) & 15.23 & 50.06 & 13.78 \\
Car & 157384 & \textbf{0.03895} (0.00022) & 0.08014 (0.00016) & 0.06685 (0.00001) & 0.16614 (0.00562) & \textbf{0.15815} (0.00457) & 0.19055 (0.00186) & 231.67 & 52.94 & 18.01 \\
Tableware & 4788 & \textbf{0.06436} (0.00009) & 0.08177 (0.00029) & 0.21082 (0.00053) & \textbf{0.18588} (0.00271)& 0.20589 (0.01163)&0.69894 (0.00030) & 5.70 & 41.63 & 13.12 \\
Ladder & 828 & \textbf{0.02767} (0.00005) & 0.08104 (0.00014) & 0.04041 (0.00010) & \textbf{0.04449} (0.00106) & 0.14899 (0.00062) & 0.06803 (0.00132) & 2.40 & 30.80 & 13.08 \\
Bull & 2418 & \textbf{0.02309} (0.00007) & 0.20242 (0.00165) & 0.04534 (0.00008) & \textbf{0.03539} (0.00165) & 0.81664 (0.00358) & 0.05535 (0.00029) & 2.91 & 56.92 & 15.08 \\
\hline
\end{tabularx}
\caption{Resulting Chamfer Distances, Hausdorff Distances, and Computation Times of Different Methods for resolution 128}
\label{table_resol128}
\end{table*}

\begin{table*}[h]
\centering
\begin{tabularx}{\textwidth}{|>{\raggedright\arraybackslash}p{2cm}|X|X|X|X||X|X|X||X|X|X|}
\hline
Name (watertight) & Num Faces & \tabularxmulticolumncentered{3}{X||}{\vspace{0pt} \large Chamfer Distance (std)} & \tabularxmulticolumncentered{3}{X||}{\vspace{0pt} \large Hausdorff Distance (std)} & \tabularxmulticolumncentered{3}{X|}{\vspace{0pt} \large Time(s)} \\ \cline{3-11} 
 &  & Diff  & 'Scan'  & DOGN  & Diff & 'Scan' & DOGN & Diff & 'Scan' & DOGN \\
\hline
Rocker-arm($\checkmark$) & 20088 & \textbf{0.01702} (0.00008) & 0.05966 (0.00015) & 0.02648 (0.00006) & \textbf{0.02691} (0.00079) & 0.10362 (0.00084) & 0.03337 (0.00053) & 198.46 & 306.67 & 103.14 \\
Beast & 64618 & \textbf{0.01502} (0.00007) & 0.06272 (0.00013) & 0.02526 (0.00002) & \textbf{0.02309} (0.00102) & 0.10318 (0.00027) & 0.08098 (0.00022) & 744.10 & 368.34 & 107.00 \\
Max Planck & 99991 & \textbf{0.02033} (0.00002) & 0.37075 (0.00201) & 0.02555 (0.00002) & 0.18064 (0.00128) & 1.08849 (0.00818) & \textbf{0.03709} (0.00180) & 1158.30 & 484.46 & 135.59 \\
Lucy($\checkmark$) & 99970 & \textbf{0.01501} (0.00003) & 0.04233 (0.00019) & 0.02463 (0.00005) & \textbf{0.02503} (0.00066) & 0.11022 (0.00372) & 0.04462 (0.00083) & 1155.86 & 345.91 & 111.01 \\
Cheburashka($\checkmark$) & 13334 & \textbf{0.01806} (0.00003) & 0.06823 (0.00002) & 0.02769 (0.00012) & \textbf{0.02977} (0.00083) & 0.16839 (0.00756) & 0.08520 (0.00111) & 130.68 & 392.88 & 121.58 \\
Happy & 98601 & \textbf{0.01726} (0.00003) & 0.06299 (0.00017) & 0.02656 (0.00004) & \textbf{0.02687} (0.00076) & 0.15838 (0.00108) & 0.03588 (0.00088) & 1140.47 & 319.96 & 114.05 \\
Teapot & 6320 & \textbf{0.01757} (0.00007) & 0.25775 (0.00125) & 0.02353 (0.00005) & 0.10390 (0.00108) & 1.01302 (0.00292) & \textbf{0.03257} (0.00040) & 61.94 & 462.17 & 114.65 \\
Bimba($\checkmark$) & 224906 & \textbf{0.02056} (0.00006) & 0.09320 (0.00020) & 0.02989 (0.00010) & \textbf{0.03400} (0.00125) & 0.11556 (0.00130) & 0.03926 (0.00236) & 3050.98 & 455.75 & 135.22 \\
Igea($\checkmark$) & 268686 & \textbf{0.02164} (0.00006) & 0.13443 (0.00021) & 0.03212 (0.00002) & 0.04057 (0.00378) & 0.10788 (0.00041) & \textbf{0.03991} (0.00023) & 3649.88 & 534.83 & 135.75 \\
Hourse($\checkmark$) & 96966 & \textbf{0.01587} (0.00002) & 0.08751 (0.00038) & 0.02600 (0.00005) & \textbf{0.02425} (0.00024) & 0.60148 (0.00601) & 0.03771 (0.00162) & 1120.32 & 418.90 & 115.71 \\
Homer($\checkmark$) & 12000 & \textbf{0.01507} (0.00001) & 0.05253 (0.00008) & 0.02510 (0.00003) & \textbf{0.02412} (0.00045) & 0.09813 (0.00144) & 0.07579 (0.00068) & 117.49 & 341.66 & 107.74 \\
Spot & 5856 & \textbf{0.01996} (0.00005) & 0.08974 (0.00039) & 0.02957 (0.00008) & \textbf{0.03278} (0.00201) & 0.11833 (0.00163) & 0.03788 (0.00242) & 57.42 & 441.26 & 120.16 \\
Armadillo($\checkmark$) & 99976 & \textbf{0.01876} (0.00003) & 0.07367 (0.00021) & 0.02858 (0.00005) & \textbf{0.03146} (0.00101) & 0.11894 (0.00291) & 0.03571 (0.00052) & 1155.40 & 356.79 & 121.72 \\
Suzanne & 968 & \textbf{0.01947} (0.00003) & 0.09038 (0.00011) & 0.03012 (0.00013) & \textbf{0.08047} (0.00191) & 0.58337 (0.00194) & 0.09545 (0.00056) & 9.79 & 430.40 & 108.23 \\
Stanford Bunny & 69451 & \textbf{0.02215} (0.00004) & 0.20425 (0.00038) & 0.02847 (0.00014) & 0.10349 (0.00072) & 0.84211 (0.00584) & \textbf{0.04049} (0.00102) & 802.02 & 479.98 & 130.56 \\
Nefertiti($\checkmark$) & 99938 & \textbf{0.01779} (0.00001) & 0.08017 (0.00040) & 0.02747 (0.00006) & \textbf{0.03064} (0.00078) & 0.10384 (0.00123) & 0.03866 (0.00489) & 1156.84 & 560.41 & 118.79 \\
Ogre & 124008 & \textbf{0.01711} (0.00006) & 0.06440 (0.00019) & 0.02906 (0.00010) & \textbf{0.07934} (0.01013) & 0.13050 (0.00250) & 0.12526 (0.00099) & 1441.33 & 375.40 & 112.09 \\
Cow($\checkmark$) & 5804 & \textbf{0.01534} (0.00004) & 0.21486 (0.00047) & 0.02528 (0.00006) & \textbf{0.02932} (0.00096) & 0.95404 (0.00273) & 0.04360 (0.00044) & 57.02 & 354.76 & 105.84 \\
Fandisk($\checkmark$) & 12946 & \textbf{0.02054} (0.00007) & 0.08235 (0.00028) & 0.03002 (0.00006) & \textbf{0.03789} (0.00251) & 0.48765 (0.00281) & 0.04194 (0.00159) & 127.07 & 321.94 & 107.73 \\
Car & 157384 & \textbf{0.03552} (0.00008) & 0.07919 (0.00037) & 0.04584 (0.00010) & 0.16396 (0.00319) & \textbf{0.16026} (0.00826) & 0.18403 (0.00240) & 1849.66 & 285.01 & 129.18 \\
Tableware & 4788 & \textbf{0.01938} (0.00005) & 0.05631 (0.00012) & 0.20433 (0.00199) & \textbf{0.04357} (0.00273)& 0.17771 (0.00911)& 0.70841 (0.00024) & 47.18 & 247.94 & 102.09 \\
Ladder & 828 & \textbf{0.01530} (0.00004) & 0.06477 (0.00031) & 0.02357 (0.00009) & \textbf{0.02224} (0.00056) & 0.12011 (0.00055) & 0.03213 (0.00047) & 16.29 & 156.47 & 101.86 \\
Bull & 2418 & \textbf{0.01802} (0.00002) & 0.21368 (0.00172) & 0.02774 (0.00003) & \textbf{0.02930} (0.00056) & 0.81471 (0.00494) & 0.03632 (0.00185) & 24.04 & 373.57 & 112.22 \\
\hline
\end{tabularx}
\caption{Resulting Chamfer Distances, Hausdorff Distances, and Computation Times of Different Methods for resolution 256}
\label{table_resol256}
\end{table*}

\end{document}